# Local Patterns Generalize Better for Novel Anomalies

Yalong Jiang, Liquan Mao

Beihang University, Beijing 100191, China

AllenYLJiang@outlook.com

**Abstract.** Video anomaly detection is a subject of great interest across industrial and academic domains due to its crucial role in computer vision applications. However, the inherent unpredictability of anomalies and the scarcity of anomaly samples present significant challenges for unsupervised learning methods. To overcome the limitations of unsupervised learning, which stem from a lack of comprehensive prior knowledge about anomalies, we propose VLAVAD (Video-Language Models Assisted Anomaly Detection). Our method employs a cross-modal pre-trained model that leverages the inferential capabilities of large language models (LLMs) in conjunction with a Selective-Prompt Adapter (SPA) for selecting semantic space. Additionally, we introduce a Sequence State Space Module (S3M) that detects temporal inconsistencies in semantic features. By mapping high-dimensional visual features to low-dimensional semantic ones, our method significantly enhances the interpretability of unsupervised anomaly detection. Our proposed approach effectively tackles the challenge of detecting elusive anomalies that are hard to discern over periods, achieving SOTA on the challenging ShanghaiTech dataset.

## 1. Introduction

Video anomaly detection (VAD) is the task of localizing from videos the events that do not match regular patterns, such as violence, accidents and other unexpected events. Nowadays, numerous platforms such as CCTVs and UAVs play an increasingly important role in surveillance. However, it is infeasible for humans to pinpoint anomalies in such an enormous amount of data among which the probability of abnormal events' existence approaches zero. Besides, the domain difference between anomalous events and normal ones leads to lack in prior knowledge about anomalies. As a result, VAD is a hot topic in weakly supervised or unsupervised learning [2,10,28,30,55,61,69,78,84,112].

Existing main-stream works [28, 43, 58] for VAD are divided into three categories. The first type of methods detect the anomalies with distinctive spatial and temporal features. Two widely used examples are prediction-based [55, 58, 61, 70] and reconstruction-based [12, 52, 62, 103] approaches. To address the issue of limited model representational capacity caused by limited training data, other methods propose to combine various feature clues [14, 28], including skeletal trajectories, appearance, motion patterns, moving directions and casual reasoning [51]. The second type of approaches apply multiple instance learning (MIL) to iterate between finding useful segments for model fine-tuning and fine-tuning models [14, 38, 53, 94, 124], using dynamic clustering to adapt models' representations to real-time observations [97, 104]. The feature spaces of some of the above approaches cannot well generalize to novel abnormal events, as is shown by Fig. 1. The third type of approaches [53] attempt to generate realistic anomalies to facilitate the construction of decision boundary between normal and abnormal samples. However, the generated anomalies are based on prior assumptions and still differ from real anomalous samples.

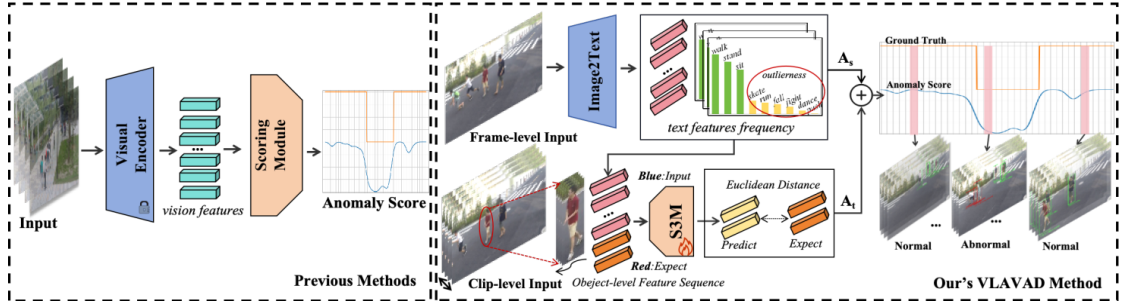

Fig. 1: Comparison between previous methods (left) and our method (right). Our purposed VLAVAD shifts from visual to semantic analysis, identifying shared attributes between normal and anomalous data while ignoring unique visual traits. Unlike traditional methods focused on specific visual cues like pose or motion, our approach is more adaptable across different scenes, facilitated by task-related semantic feature selection. Additionally, we introduce the Sequence State Space Module (S3M) to learn the temporal correlation of normal samples, thereby detecting anomalies that deviate from the normal temporal pattern.

Our proposed method, VLAVAD, eliminates the need for collecting and labeling anomalous data, making it suitable for real-world applications. By utilizing Selective Prompt Adapter(SPA) and employing a lightweight S3M trained on normal data, our approach effectively harnesses the deep semantic information in images, allowing for precise and interpretable spatiotemporal localization of anomaly events. The method has been successfully validated across multiple datasets, showcasing its cost-effective transferability and superior performance.

In summary, our contributions can be summarized as follows:

- We present an unsupervised video anomaly detection framework called VLAVAD, which utilizes semantic features rather than visual features for anomaly detection. This framework capitalizes on the comprehension and reasoning skills of pretrained Visual Language model to enhance performance in VAD. Consequently, our method expands the anomaly detection from a particular dimension to open-world.

- We introduce the pioneering use of the Sequence State Space Module (S3M) to tackle temporal variation in anomaly detection, further mitigating the limitation of single frame anomaly assessment that overlooks time-related anomalies.

- Our method allows for cost-effective universal anomaly event discrimination across scenes,

achieving a 2.7% improvement in performance on the challenging cross-scene, cross-category Shanghaitech dataset. We also validate the superiority of our approach across multiple datasets.

## 2. Related Work

### 2.1. Unsupervised Video Anomaly Detection

Due to the imbalanced nature of surveillance videos, most available training datasets are without anomaly annotations because it is expensive to label [17, 42, 50]. Reconstruction-based approaches [1, 6, 24, 30, 41, 79, 103] produce increased error when encountering irregular spatio-temporal features [75] that do not reside in training data. For instance, [64, 108] improved model structures for better reconstruction. [27, 30] augmented encoders to improve the sensitivity of reconstruction error to anomalies. [12, 63, 85, 107] integrated appearances, motion features, audio features and rule-based features [19]. [34] reconstructed images with a probabilistic decision model. [111] distinguished good and bad quality reconstructions to improve stability. Prediction-based methods such as [48, 58, 59, 66, 67, 114] evaluated the divergence in normal and abnormal temporal dependencies, leveraging latent spaces [118] or hybrid attention [117].

To better distinguish anomalies under ambiguous cases, [52, 55, 61, 70] combined prediction with reconstruction and built a pool of features for encoding normal dynamics. [60, 81, 98] studied the distribution over normal patterns and proposed novel deep features [4] to separate anomalies from normal samples. Similarly, [101] proposed denoising diffusion modules to learn the distribution of normal events. [26] exploited the enhanced mode coverage capabilities of diffusive probabilistic models. To distinguish anomalous patterns with better representations, [101] proposed contrastive and snippet-level anomalous attention. [26] introduced pyramid deformation module and localization mechanism to enhance the power of reconstruction. [72] leveraged CRFs to learn the dependencies across frames. To fully exploit task-relevant features, [86] combined interpolation with extrapolation for prediction. [95] proposed a self-supervised learning scheme with discriminative DNNs. [83] proposed to sequentially learn multiple pretext tasks to enhance anomaly detection. Although remarkable improvements have been achieved, some of the representations cannot well represent unseen abnormal patterns.

### 2.2. Weakly Supervised Anomaly Detection

Multi-instance learning (MIL) takes videos as bags and snippets as instances, transforming video-level labels to instance-level supervision [25]. The methods iteratively locate abnormal segments and fine-tune models using the segments. To collect abnormal segments, inter-sample similarities are evaluated [35, 56]. For instance, clustering-based approaches measured the similarity in spatio-temporal embeddings [12, 18, 62, 65]. proposed a probabilistic framework for categorizing actions. [89] built graphical representations connecting different objects. [14] integrated collective properties in measuring similarities. Then the anomalous segments which are dissimilar to normal ones [115] function in fine-tuning. [80] performed dynamic non-parametric clustering and exposed the model to potentially positive instances. To improve robustness and efficiency, [119] proposed to interpret the vulnerability of MIL. [99] introduced casual relations to enhance MIL [91]. [102] proposed binary network augmentation strategy to improve detection performance.

## 2.3. Methods with Data Augmentation

To generate pseudo abnormal samples as supervision in fine-tuning models, methods such as [7, 36, 46, 47, 53] proposed pseudo abnormal snippet synthesizers which are trained on normal samples [105]. [110] employed a generator which was not fully trained to create abnormal samples as supervision. [13] proposed to generate class balanced supplementary training data with a conditional GAN. [45] focused on infrequent normal samples during generation, harnessing novel sampling strategies. Besides frame-level analysis [112], human-level approaches [35, 58, 90] provide more fine-grained analysis. Similarly, [20] identified the outliers as positive samples by assigning anomaly scores to objects. [3] introduces a new dataset with diverse anomalies. However, the generated samples are based on normal patterns and still differ from real-world anomalies.

## 2.4. Methods Exploring the Representation of Unseen Categories

To adapt model representations and work under changing anomalies, meta learning-based methods such as [55, 70], transfer-learning based approaches [20, 71] and self-supervised approaches [16, 69] introduced adjustable feature representations to adapt to new domains. Attention-based methods such as [32, 40, 57, 87, 88] attended to domain-invariant features in addressing unseen samples while reducing background influences. To better align with anomaly detection, [28] integrated multiple sub-tasks, including moving direction prediction, appearance consistency evaluation and object classification. [123] introduced multi-level graphs for representing videos and maximizing the margins between normal and abnormal ones. Differently, our proposed approach locates the local patterns which generalize to unseen categories of events. The patterns are learned from image-text alignment. Besides spatial patterns, the dynamics of local patterns are modeled with state machines [31] which are embedded with motion components.

## 2.5. Prompting Methods

Prompt-based approaches have been widely used in anomaly detection [22, 53, 53, 81]. Different from the approaches which leverage complex backbones, the proposed approach combines a Resnet-18 based backbone with an image-text alignment module for obtaining language-informative local patterns.

## 3. Our Method

Our main objective is to develop an unsupervised learning methodology to effectively handle scenarios with unpredictable and unobtainable anomalous data samples. Our approach involves transitioning from vision to semantic features, identifying common attributes between normal and anomalous data in the semantic space while excluding non-shared visual features. In contrast to conventional methods that heavily rely on specific aspects of visual features such as pose or optical flow data, our approach offers a significant advantage in its seamless adaptability across diverse cross-scene datasets, facilitated by the incorporation of a Prompt Adapter. Additionally, we introduce the Sequence State Space Module (S3M) to detect temporal variations in semantics, complementing single-frame detection results and addressing the limitation of underutilizing temporal information in anomaly detection.

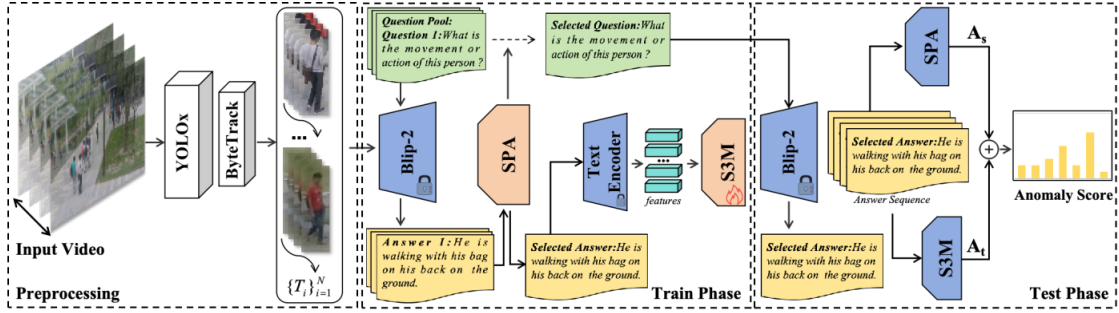

Fig. 2: Overview of our purposed VLAVAD. In the preprocessing stage, object-level sequences $\{T_i\}_{i=1}^N$ are obtained through detection and tracking. During training, the Selective Prompt Adapter (SPA) selects the most suitable prompt from the prompt pool to describe the dataset scene samples. Subsequently, the Sequence State Space Module (S3M) takes clip-level semantic features $E(t)$ as input and is trained using Mean Squared Error(MSE) loss between the predicted feature output and the expected feature to learn the deviations in temporal patterns. During testing, we utilize the prompt selected by SPA from the training set to generate the answer sequence. We then calculate As and At, which represent the static caption anomaly score and time inconsistency anomaly score, respectively.

### 3.1. Obtain Multi-object Trajectories

Our Anomaly Detection Architecture receives a series of object-level temporal image sequences for input. To achieve object detection, we employ a pre-trained YOLOx network. Additionally, we utilize the ByteTrack algorithm for object tracking to train the S3M. Consequently, we acquire object-level trace trajectories $T = \{O_i\}_{i=f_{begin}}^{f_{end}}$. Finally, we obtain an object-level trajectories set $\{T_i\}_{i=1}^N$, where $N$ is the total number of objects detected in the video, which facilitates the segmentation of each object into clips during both training and testing phases.

### 3.2. Algorithm Description

Illustrated in the right half of Fig. 2, our network comprises three components. The first component, the Selective Prompt Adapter, employs the frequency distribution of the output of LLM to compute anomaly scores for individual objects detected within a single frame. It selects the most salient score among multiple objects within the same frame and designates it as the anomaly score for that frame, denoted as $A_k = \max_{i=1}^n (A_{O_i})$, where $A_k$ represents the anomaly score for the $k$-th frame and $A_{O_i}$ represents the anomaly score for the $i$-th object within that frame. The second component, the Sequence State Space Module (S3M), takes as input the object-level text embedding sequence generated by VLM. It undergoes unsupervised training solely on the normal samples within the training set and computes anomaly scores based on the temporal inconsistency of features during the test phase. Finally, we integrate the static anomaly scores with the dynamic ones and apply Gaussian smoothing to obtain the final score.

## 4. Experiments

### 4.1. Dataset and Metric

To demonstrate the effectiveness of the proposed framework, we conduct experiments on datasets: ShanghaiTech [48], CUHK Avenue [54]. The training sets of ShanghaiTech, Avenue and UCSD Ped2 contain only normal events and abnormal behaviors reside in test data.

**ShanghaiTech** dataset contains 330 training videos and 107 test videos with 130 abnormal events. Typical anomalies include fighting, running, cycling and so on. Among the two versions of ShanghaiTech dataset [48] and [38, 120], the latter [113, 124] includes abnormal behaviors in both training set and test set. As our approach is unsupervised, we use the first version. HR-ShanghaiTech includes only human-related video, six non-human test videos are neglected [33].

**CUHK Avenue** dataset involves 16 and 21 video clips for training and test, respectively. The dataset covers abnormal movements and moving directions. In HR-Avenue, non-human anomalies are ignored [66].

**Evaluation Metrics** Following previous literatures [65], Area under Curve (AUC, \%) is adopted for evaluation, it is computed by continuously changing the threshold for anomaly detection before conducting integration. A higher AUC value indicates better performance. Different from other datasets, the accuracy on XD-Violence dataset is measured by precision-recall curve and the corresponding Average Precision (AP, %) [68].

### 4.2. Implementation Details

As is shown by Fig. 2, an off-the-shelf Region Proposal Network (RPN) [77] is leveraged. For each detected object, if it is overlapped with neighboring bounding boxes, then we merge it with neighbors. If the merged bounding box is not squared, we enlarge its regions along the short side to include more background contexts. Each bounding box is resized to $224 \times 224$ before being fed into backbones. The output from backbone $H_{i,b}^{I}(t) \in \mathbb{R}^{S_d \times V_d}$ is projected by the Image-attention Module in ITAM to $F_{i,b}^{I}(t) \in \mathbb{R}^{N_q \times E_d}$ which is the output of the spatial part of the framework. $S_d = 257$, $V_d = 1408$, $N_q = 32$, $E_d = 256$. The detailed structures will be illustrated in supplementary materials.

The backbones in Branch 1 and Branch 2 have Resnet-18 structures with only first two stages kept, producing a $28 \times 28 \times 512$ tensor for each image. The tensor is resized to $16 \times 16 \times 1408$. Class token [21] with size $V_d$ for each input image is concatenated with backbones' resized outputs, producing $H_{i,b}^{I}(t) \in \mathbb{R}^{S_d \times V_d}$. Branch 1 and Branch 2 are trained independently. We firstly train the spatial part to extract features $\mathbf{v}_i(t)$, then freeze the spatial part and train SMM to predict future features based on past ones.

**Hyperparameters for training Spatial Part** The learning rate schedule is Linear Warmup With Cosine Annealing. The warmup learning rate is $10^{-6}$ which increases to initial learning rate $10^{-4}$ and then decreases to minimum learning rate $10^{-5}$ in a cosine annealing learning rate schedule. The warmup stage lasts for 5000 steps. The batch size for training is 120.

**Hyperparameters for training Temporal Part** The learnable weights in SMM include those in the Object Feature Encoder (OFE) with $N_q E_d$ input channels and $O=64$ output channels and those in Object Feature Decoder (OFD) with $O$ input channels and $N_q E_d$ output channels. The OFE and OFD are linear layers. Besides, the weights in $C \in \mathbb{R}^{O \times O}$ are learnable. All weights are initialized according to distribution $N(0, 0.02)$. Training lasts for 20 epoches, initial learning rate is $5 \times 10^{-5}$ with learning rate decay 0.99. The cross attention layer for combining the outputs from branches is trained together with SMM.

Implementations are based on Pytorch platform [73]. Experiments are conducted on one NVIDIA A100 GPU. On the datasets [88, 100, 3] where training data includes anomalies, the procedures for training the spatial part of the framework remain the same. For instance, CLIP model [74] is also leveraged in generating labels. The spatial part learns to produce different spatial arrangements of local patterns in different events. The temporal part of the framework only learns on the training videos without anomalies.

## 4.3. Comparisons with Related Methods

Table 1: Performance (AUC, %) on the ShanghaiTech, CUHK Avenue, Ubnormal and UCSD Ped2. Micro-AUC [76] is evaluated.

| Pub. Year | Methods | UCSD Ped2 | Avenue | ShanghaiTech |
|---|---|---|---|---|
| 2018 before | MPPC+SFA[40] | 61.3% | - | - |
| | Conv-AE[22] | 90.0% | 70.2% | - |
| | ConvLSTM-AE[38] | 88.1% | 77.0% | - |
| | StackRNN[39] | 92.21% | 81.71% | 68.0% |
| | Frame-Pred[35] | 95.4% | 85.1% | 72.8% |
| 2019 | Mem-AE[20] | 94.1% | 83.3% | 71.2% |
| | AnoPCN[64] | 96.8% | 86.2% | 73.6% |
| | Deep-OC[61] | 96.9% | 86.6% | – |
| 2020 | ClusterAE[8] | 96.5% | 86.0% | 73.3% |
| | IPR[54] | 96.20% | 83.70% | 71.50% |
| | MNAD-Recon[48] | 90.2% | 82.8% | 69.8% |
| 2021 | CT-D2GAN[15] | 97.2% | 85.9% | 77.7% |
| | LNRA[3] | 96.5% | 84.7% | 76.0% |
| 2022 | ARAE[29] | 97.4% | 86.7% | 73.6% |
| | CR-BPN [9] | 98.3% | 90.3% | 78.1% |
| 2023 | MGME [69] | 97.8% | 87.6% | 73.5% |
| | SPTD[28] | - | - | 84.5% |
| | OFR-E[24] | 97.7% | 89.7% | 75.8% |
| 2024 | STM[30] | 97.0% | 87.7% | 76.1% |
| | CR-KR[6] | 97.1% | **90.8%** | 83.7% |
| | **Ours** | **99.0%** | 87.6% | **87.2%** |

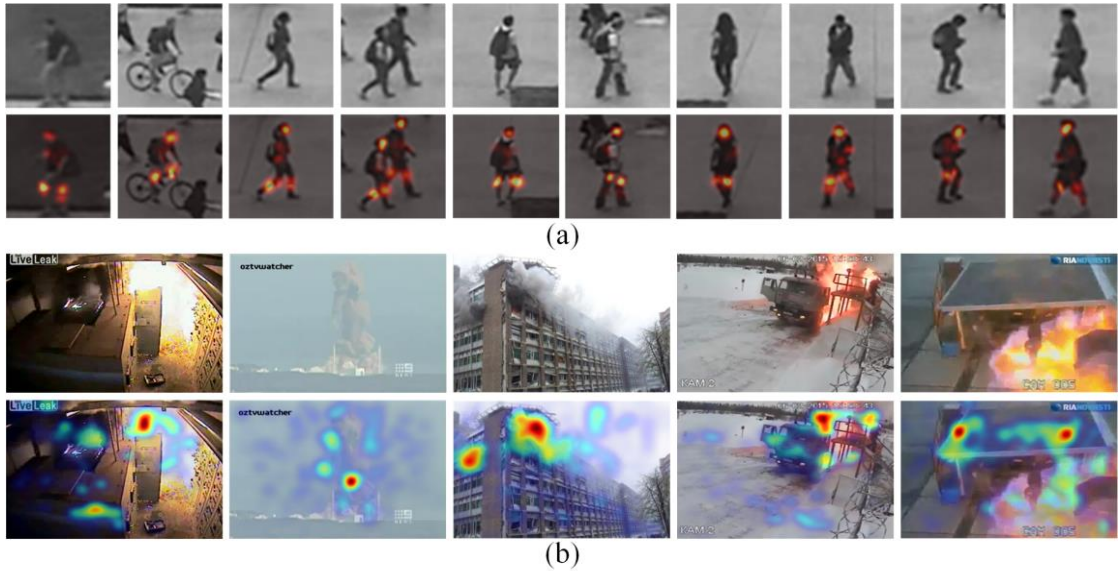

**Fig. 4:** Heatmaps of local patterns. (a) Low-resolution humans. (b) Abnormal objects whose spatial distributions of local patterns differ from those of normal objects.

The proposed approach is compared with existing ones in detecting different types of anomalies, as is shown in Table 1. Then the effectiveness is verified on different types of objects.

### 4.4. Subjective Results

The local patterns in low-resolution videos can also be located. Example data are from UCSD Ped2 dataset where the resolution of a single human is about $30 \times 30$ pixels. It can be seen from Fig. 4 that spatial patterns can be accurately located regardless of low resolutions. The heatmaps are obtained with [82] based on the data in $H_{i,b}^{I}(t)$.

## 5. Conclusion

Previous efforts in video anomaly detection have typically relied on visual representations, which has limited the ability to generalize across diverse situations. For instance, behaviors that are considered typical in one context may be deemed anomalous in another. Our method addresses this challenge by employing the Selective Prompt Adapter (SPA) to enable a pretrained VLMs to perform cross-scenario, interpretable anomaly detection more effectively. The advancement of cross-modal large models, as well as the progress in cross-modal matching models and Language Language Models (LLMs), has made it possible to extend this technique to enhance the interpretability and generalization of VAD.